\documentclass[12pt]{article}    

\usepackage{times,fullpage}
\usepackage{graphicx}     
\usepackage{amsmath}
\usepackage{txfonts}

\newtheorem{THEOREM}{Theorem}
\newenvironment{theorem}{\begin{THEOREM} \hspace{-.85em} {\bf :} }%
                        {\end{THEOREM}}

\newtheorem{LEMMA}[THEOREM]{Lemma}
\newenvironment{lemma}{\begin{LEMMA} \hspace{-.85em} {\bf :} }%
                      {\end{LEMMA}}
\newtheorem{COROLLARY}[THEOREM]{Corollary}
                          {\end{COROLLARY}}
\newtheorem{PROPOSITION}[THEOREM]{Proposition}
\newenvironment{proposition}{\begin{PROPOSITION} \hspace{-.85em} {\bf :} }%
                            {\end{PROPOSITION}}
\newtheorem{DEFINITION}[THEOREM]{Definition}
                            {\end{DEFINITION}}
\newtheorem{CLAIM}[THEOREM]{Claim}
                            {\end{CLAIM}}
\newtheorem{EXAMPLE}[THEOREM]{Example}
                            {\end{EXAMPLE}}
\newtheorem{REMARK}[THEOREM]{Remark}
                            {\end{REMARK}}
							\newtheorem{NOTATION}[THEOREM]{Notation}
							                            {\end{NOTATION}}
\newenvironment{proof}{\noindent {\bf Proof:} \hspace{.677em}}%
                     {}

\DeclareMathAlphabet{\mathitbf}{OML}{cmm}{b}{it}

\newcommand{\blemma}{\begin{lemma}}

\newcommand{\elemma}{\end{lemma}}

\newcommand{\bthm}{\begin{theorem}}
\newcommand{\ethm}{\end{theorem}}
\newcommand{\bprf}{\begin{proof}}
\newcommand{\eprf}{\end{proof}}
\newcommand{\bpro}{\begin{proposition}}
\newcommand{\epro}{\end{proposition}}
\newcommand{\bi}{\begin{itemize}}
\newcommand{\ei}{\end{itemize}}
\newcommand{\be}{\begin{enumerate}}
\newcommand{\ee}{\end{enumerate}}
\newcommand{\beq}{\begin{equation}}
\newcommand{\eeq}{\end{equation}}
\newcommand{\bcase}{\begin{cases}}
\newcommand{\ecase}{\end{cases}}

\begin{document}

	
\title{The Quest for Interpretable and Responsible \\ Artificial Intelligence\thanks{This is a slightly edited version of an article to appear in  {\it The Biochemist}, Portland Press. References and material for further reading can be found at the end of the article.}} 

\author{{\bf Vaishak Belle} \\ 
University of Edinburgh \& Alan Turing Institute \\ 
{\small vaishak@ed.ac.uk} 
}

\date{}

\maketitle

\begin{abstract} \it Artificial Intelligence (AI) provides many opportunities to improve private and public life. Discovering patterns and structures in large troves of data in an automated manner is a core component of data science, and currently drives applications in computational biology, finance, law and robotics. However, such a highly positive impact is coupled with significant challenges: How do we understand the decisions suggested by these systems in order that we can trust them? How can they be held accountable for those decisions? 
	
	In this short survey, we cover some of the motivations and trends in the area that attempt to address such questions.
	
\end{abstract}

\section{Introduction} 
\label{sec:introduction}

Since at least the 80s, AI had enjoyed considerable success in a wide variety of applications, such as logistics and automated diagnosis. In most cases, knowledge about the problem was expressed in a symbolic framework by an expert, and a search procedure was used to identify a solution. In contrast, the last two decades have seen the explosive growth of methods that learn models directly from data. Among other things, the availability of large repositories of suitable training data, and the increase in computer processing power have been key factors. The popularity of machine learning (ML) lies in the fact that humans have a propensity to model problem domains in a rigid and deterministic way. Such models may fail to identify hidden patterns, or otherwise deal with the randomness found in nature.  By learning models from data, some of these pitfalls may be avoided. Recently, for example, the computer program AlphaGo beat professional human players at the ancient strategy game Go; the program was trained on 30 million positions. 

Inspired by such successes, there is a widespread trend to leverage state-of-the-art ML techniques for broad applications. These include healthcare, law, finance, robotics and self-driving cars. However, many of these techniques are virtual blackboxes, that is, their decision logic is not understandable to us. This is very problematic for a number of reasons, which are perhaps best illustrated by the two cases below.

\begin{center}
	\includegraphics[width=.8\textwidth]{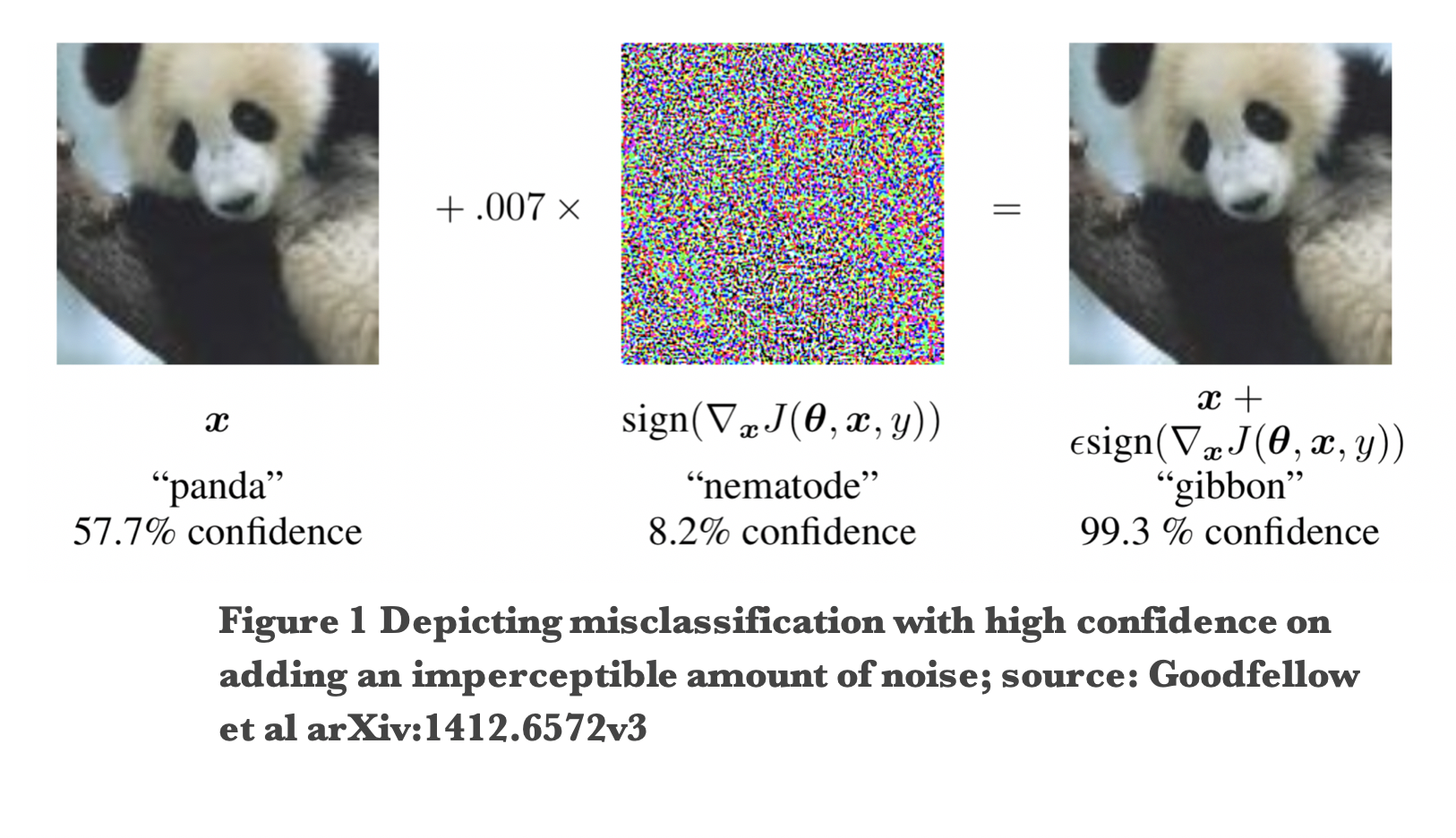}
\end{center}

In a now widely cited experiment, a neural network can be guided to misclassify, with high confidence, a panda as a gibbon when an imperceptible amount of “noise” is added to the data. In domains where surprises are infrequent, such techniques nonetheless have very high accuracy across standard test images, and so the concern may seem needlessly pedantic. But in self-driving cars, where encountered images may have massive variability under diverse lighting conditions, misclassification and confusion could lead to serious passenger/pedestrian injuries. As an extreme example, imagine if the presence of background texture could result in a child being mistakenly identified as a road surface marking. To invoke a biomedical example, a recent prediction model for inferring the risk of death for patients who developed pneumonia suggested, counterintuitively, that asthmatics are less likely to die from pneumonia. The reason? Existing policy recommends aggressive and immediate treatment of asthmatics with pneumonia. Rectifying and revising the model is, of course, always possible, and clinicians would not uncritically accept a machine's opinion over their clinical experience. However, the take home message from both cases is that we can never anticipate every possibility, and machine-generated patterns need to be scrutinized and tested, not applied unthinkingly. Interpretability can also be useful in less extreme situations, such as approving credit card applications. Here, for legal reasons, we may need to justify a decision, and moreover, prove that no discriminatory action was practiced.


\section{Enabling Interpretability} 
\label{sec:enabling_interpretability}

Interpretability is hard to define precisely. But perhaps we can draw an analogy to a dynamical system. When considering a moving object, we likely understand what caused the object to move, but also what can cause it to stop. Such a system can be interrogated to investigate the impact of changes: how much distance would the object have moved if it was heavier? Or the force was greater? Or if the floor was smoother? And so on. The system can also be specified concisely and so we are able to study the laws governing the movement. 

Expecting learning methods to demonstrate this type of logic is challenging, perhaps impossible, but can be seen to essentially relate to ideas from the early days of AI that required computation and outputs to be understandable. Indeed, for enabling interpretability, learning methods increasingly draw technical concepts from classical AI.

\begin{center}
	\includegraphics[width=.8\textwidth]{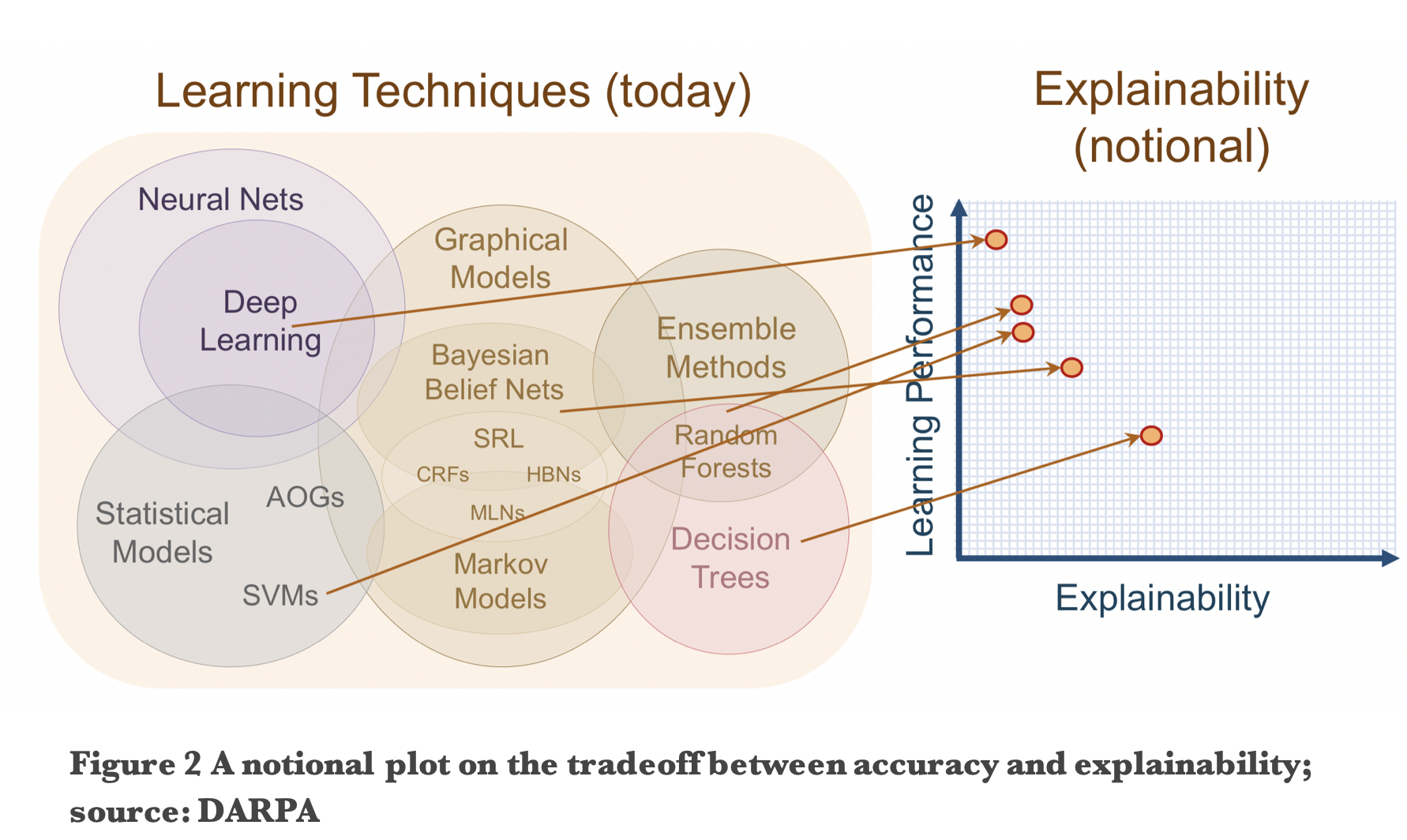}
\end{center}

Conversely, are there reasons for \textit{not} insisting on interpretability? One argument is that blackbox methods are more accurate, a view expressed by the USA's Defense Advanced Research Projects Agency (DARPA) in their notional plot of how ML methods compare in terms of accuracy vs interpretability (Figure 2). This plot has since been criticized for being ill-defined, and in case where blackbox methods had performed better, interpretable methods seem to be closing the gap. A second argument is that people often provide ad hoc rationalizations to their actions. Whilst this may be true (and acceptable) in social and personal situations, it is surely not something we would wish to aspire to when it comes to understanding the physical/chemical/biological principles of the universe. 
A third argument is that all models are interpretable, all we need to do is read the computer code; however, even if the architecture and training regime are transparent, it tells us very little about the decision logic of the system learned over many training epochs and ingenious feature engineering. We will now discuss various strategies for addressing interpretability, not all mutually exclusive and offering different strengths. 

The first natural approach is that of {\it interpretable model specification}. This arises in the context of viewing tasks such as classification, prediction, and labeling as a probabilistic computation, by appealing to the Bayes’ law. Moreover, paradigms such as probabilistic programming have recently emerged to simplify the process of specifying a statistical model and learning its probabilities. A model for how infection spreads on contact in a population, for example, can be expressed using only 3 lines of code; the probability of an individual getting in contact with other members of the population can then be learned from data.
Like traditional computer programs, re-usable computational tasks can be written in an isolated manner, and then referenced in a more complex instruction. Thus, challenging applications can be written in a principled way. Executing such a program would provide a justification for an outcome, which is interpretable to the extent that it reflects how the probabilities were obtained and the prediction calculated. 

A second approach is to learn \textit{interpretable classifiers}. Standard ML methods such as decision trees and linear regression were, in fact, early examples of such an approach. Other more recent methods attempt to induce programmatic/logical structures from data. The long-term vision is that these structures would recover the causal or generative process behind the data. One caveat, however, is whether the vocabulary of the classifier coincides with that of the user’s intuitive understanding of the problem domain: classifier features that are too granular or too vague will likely be hard to understand. We then may need to “map” classifier features to terms that the user would find helpful and interpretable.

Since blackbox classifiers may turn out to be large and unwieldy, a third approach is to emphasize and support \textit{interactive querying}. For example, if the model for the infection spread was obtained instead by means of a complex pipeline expressed using many lines of code, it suffices that the pathologist would be able to ask observational queries such as: how does the infection rate differ from one social network to another?, or if we sampled friends from one network and found them infected, what does it say about the entire population? It may, moreover, be advantageous to move beyond purely statistical associations. In the case of EU’s General Data Protection Regulation, an argument has been made recently for supporting counterfactual querying. For example, when confronted with the decision that one’s credit card application was denied, the model would suggest how certain types of changes to the applicant’s circumstance would get the decision reversed. 

In general, counterfactual querying is a type of causal inference, which requires us to specify not only the probabilistic dependencies between variables but also the mechanism that determines the values of variables. This can be hard to identify in the absence of experts. 
It is also worth remarking that most ML models inherently appeal to statistical associations and so one could accuse all of them of not possessing a decision logic per se. Thus, empowering statistical methods with causal, analogical and logical structures will perhaps be needed eventually to generate explanations with a clear decision logic. 

For blackbox classifiers, a fourth approach attempts to construct summaries by inspecting, augmenting or otherwise approximating the decision logic of the model. In particular, proposals on \textit{post hoc interpretability} sample very close to the region of interest, and then an interpretable classifier is generated that serves as a local approximation. However, such approaches have since been criticized because it is not always clear that these approximations are faithful to how the model actually works. 

If we reconsider the dynamical system analogy, there may be a distinction to be made between having an interpretable model, where an expert is in a position to understand the nature of the computation and the output in principle, versus obtaining an explanation. In this sense, an extremely large decision tree or induced program may seem impenetrable to a lay person. How to enable such explanations is a point of debate and part of ongoing research. While textual and visual rationalizations are always possible on an ad hoc basis, social scientists and philosophers have argued that explanations need to be contrastive (why this and not that), minimal (clarifying the relevant entities) and social (taking the asker’s knowledge into account). Thus, approaches that leverage causal inference while positing a mental model of the user are likely to be successful for generating explanations at this level of sophistication.

\section{Towards Responsible AI} 
\label{sec:towards_responsible_ai}

Enabling interpretability is one concrete way for the responsible deployment of AI systems, but it is far from the only one. When we consider that AI-based predictions act on people, we need to champion responsible deployment more generally. As a consequence of datasets reflecting decades of historical and cultural biases, there is growing alarm that ML systems continue to manifest inherited prejudices against certain groups. In one recent case of considerable notoriety, Pro-Publica, a US-based organization specializing in not-for-profit journalism, published an article suggesting that an algorithm widely used to predict the probability of re-offense in criminals was biased against African-American offenders. While the analysis of that article has since been criticized for misunderstanding the original algorithm’s risk allocation measures, there is still the potential for injustices to arise as a consequence of data hinging on sensitive factors that may be difficult to identify.  In fact, the data need not even be biased against certain groups; it suffices that the data is representative of one group, say Caucasians, but is nonetheless deployed in a diverse world, leading to invalid or offensive predictions. 

A growing literature now attempts to force classifiers to be \textit{fair}, in the sense of ensuring that predictions do not change based on sensitive factors. For example, the notion of predictive parity says that a classifier is considered fair if it is equally likely to make a positive identification regardless of the value of a sensitive factor, such as gender. Not surprisingly, at this early stage, there is considerable debate still on which definition correctly formalizes equality and demographic parity, especially owing to the fact that some of these definitions are mutually inconsistent, and how that should be implemented. 

\begin{center}
	\includegraphics[width=.8\textwidth]{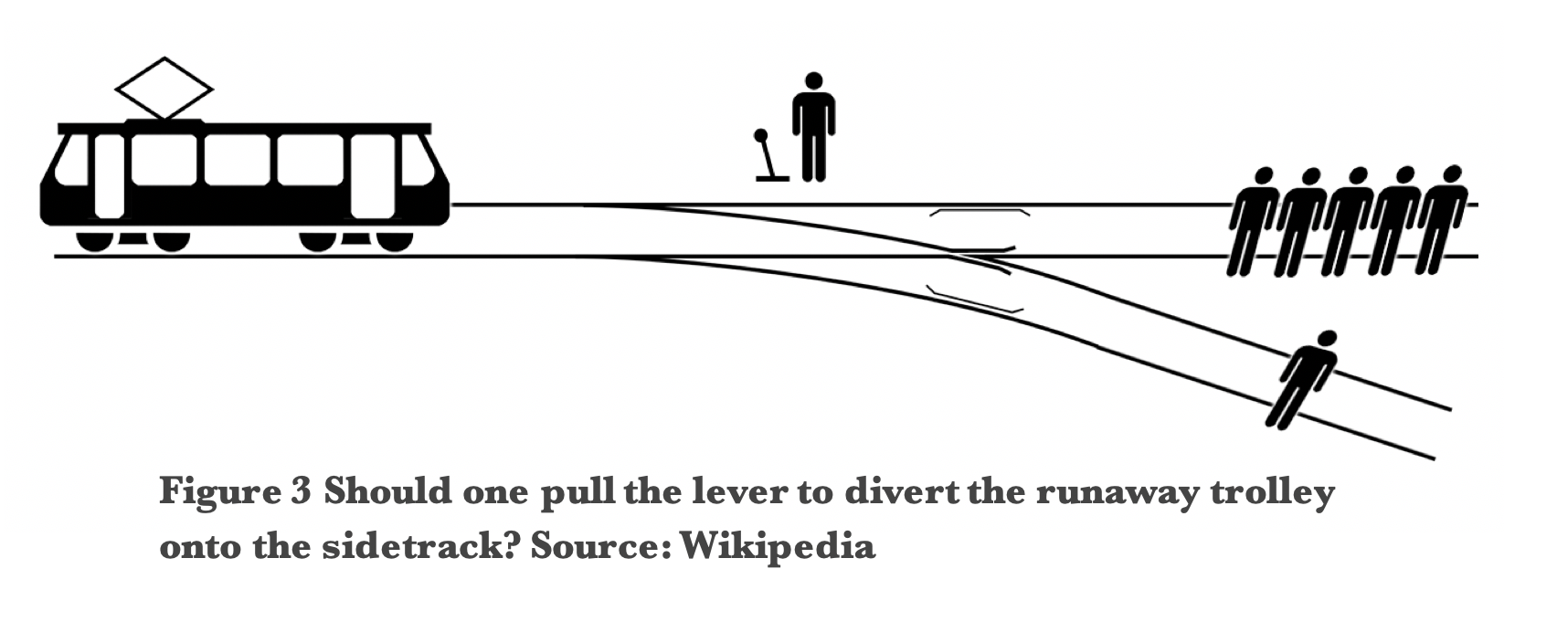}
\end{center}

Fairness, is however, one part of a larger picture on making AI responsible. \textit{Value alignment} is a notion that attempts to ensure that a system’s objectives aligns with human values. But this can be challenging to realize because human society can disagree significantly on which values matter. Our views on, for example, the ethical treatment of peoples and other animals, has changed radically in the last century alone, and classical thought experiments such as the \textit{trolley problem} (see Figure 3) remain notorious, unresolved and controversial. Even if we agreed on such values, there are long-standing concerns about translating ethical principles into a framework based on numeric representations, denoting, say, a consequentialist stance (i.e. one based on outcomes). Nonetheless, progress is perhaps only possible by stipulating models and hypotheses, and despite its delicate nature, numeric models are not uncommon in social contexts, such as the use of quality-adjusted life years within healthcare and the value of life within economic and insurance policies. Frameworks have thus been emerging to provide computational mechanisms for representing and reasoning about ethical decisions. The crucial step when allowing systems to reason about their action choices is to postulate the utilities and costs of those choices, which can determine the more responsible thing to do. In the context of, say, lung cancer staging, such costs can represent a decision strategy: a thoractomy is the usual treatment unless the patient has mediastinal metastases, in which case a thoractomy will not result in greater life expectancy than the lower risk option of radiation therapy, which might then be the preferred treatment. In the trolley problem, where a runaway trolley on a path to kill 5 people can be diverted to a side track that kills one person, costs can denote a composite of emotive and analytic traits of the decision maker. In empirical studies, major differences have been observed across cultures when considering situations that variably introduce elderly people, children, pets, and family members in the main and side tracks. Thus, one could view such frameworks as a means to provide, on the one hand, a higher degree of autonomy for reasoning about the consequences of actions, and on the other, personalised models learnt from data that encode moral preferences. Contextualizing these preferences is very important, however: we would consider it immoral to harvest a healthy person’s organs to save five people. Moreover, responsibility and interpretability are interlinked: to prove a claim that one was unbiased and morally right, say in the court of law, an explanation is needed as to why the disputed action was taken in the first place. 

Ultimately, the goals of the emerging trends in interpretable and responsible AI can be seen as an attempt to enable technology that benefits human society. This places serious demands on the accessibility and justifiability of that technology, which effects and is affected by political and public discourse on the use of AI. The scientific endeavor is simply the start of a dialogue, and mathematical advances allow us to be increasingly concrete in the applications considered and the broader impacts thereof.

\section*{Further reading} 
\label{sec:references}

\begin{itemize}
	\item Goodfellow, I.J., Shlens, J. and Szegedy, C. (2015) Explaining and harnessing adversarial examples. ICLR arXiv:1412.6572v3
\item Gunning, D. (2019) DARPA's explainable artificial intelligence (XAI) program. IUI
\item  Weld, D.S. and Bansal, G. (2019) The challenge of crafting intelligible intelligence.
Commun. ACM 62, 70–79
\item Pearl, J. (2019) The seven tools of causal inference, with reflections on machine
learning. Commun. ACM 62, 54–60
\item Russell, S.J., Dewey, D. and Tegmark, M. (2015) Research priorities for robust and
beneficial artificial intelligence. AI Magazine 36
\item Miller, T. (2019) Explanation in artificial intelligence: insights from the social sciences.
Artif. Intell. 267, 1–38
\item Caruana, R., Lou, Y., Gehrke, J., Koch, P., Sturm, M. and Elhadad, N. (2015) Intelligible
models for healthcare: predicting pneumonia risk and hospital 30-day readmission. KDD \item Wachter, S., Mittelstadt, B.D. and Russell, C. (2018) Counterfactual explanations without
opening the black box: automated decisions and the GDPR. Harvard Journal of Law \&
Technology, 31, 841–887
\item De Raedt, L., Kersting, K., Natarajan, S. and Poole, D. (2016) Statistical relational artificial
intelligence: logic, probability, and computation. Synthesis Lectures on Artificial
Intelligence and Machine Learning, Morgan \& Claypool Publishers
\item Hemment, D., Aylett, R., Belle, V., D. Murray-Rust, E. Luger, J. Hillston, M. Rovatsos and F. Broz.  (2019) Experiential AI. AI Matters 5, 25–31
\item Belle, V. (2017) Logic meets probability: towards explainable AI systems for uncertain
worlds. IJCAI
\item Hammond, L. and Belle, V. (2019) Deep tractable probabilistic models for moral
responsibility. NeurIPS Workshop on Knowledge Representation \& Reasoning Meets Machine Learning. CoRR abs/1810.03736

\item Rudin, C. (2018) Please stop explaining black box models for high stakes decisions. NeurIPS Workshop on Critiquing and Correcting Trends in Machine Learning. CoRR
abs/1811.10154
\end{itemize}




\bibliographystyle{abbrv}
\bibliography{group}
\end{document}